\documentclass[runningheads]{llncs}
\usepackage{graphicx}
\usepackage[ruled,vlined]{algorithm2e}
\usepackage{multirow}
\usepackage[T1]{fontenc}
\usepackage{mwe}    
\usepackage{subfig}
\usepackage{pdfpages}
\usepackage{graphicx}
\usepackage{multicol}
\usepackage{multirow}
\usepackage{footnote}
\usepackage{comment}
\usepackage{parskip}
\usepackage{amsmath}
\usepackage{amssymb}
\usepackage{algorithmic}
\usepackage{url}
\begin{document}
\setlength{\parskip}{0.1pt}
\title{
A Structural-Clustering Based Active Learning for Graph Neural Networks
}
%
%

\author{Ricky Maulana Fajri\inst{1} \and
Yulong Pei\inst{1} \and
Lu Yin\inst{1,2}\and
Mykola Pechenizkiy \inst{1}}
\authorrunning{Fajri et al.}
%
\institute{Eindhoven University of Technology, The Netherlands 
\email{\{r.m.fajri,y.pei.1,mykola.pechenizkiy\}@tue.nl}
\and The University of Aberdeen, UK\\
\email{lu.yin@abdn.ac.uk}
}
\maketitle              
\begin{abstract}
In active learning for graph-structured data, Graph Neural Networks (GNNs) have shown effectiveness. However, a common challenge in these applications is the underutilization of crucial structural information. To address this problem, we propose the \textbf{S}tructural-Clustering \textbf{P}ageRank method for improved \textbf{A}ctive learning (SPA) specifically designed for graph-structured data. SPA integrates community detection using the SCAN algorithm with the PageRank scoring method for efficient and informative sample selection. SPA prioritizes nodes that are not only informative but also central in structure. Through extensive experiments, SPA demonstrates higher accuracy and macro-F1 score over existing methods across different annotation budgets and achieves significant reductions in query time. In addition, the proposed method only adds two hyperparameters, $\epsilon$ and $\mu$ in the algorithm to finely tune the balance between structural learning and node selection. This simplicity is a key advantage in active learning scenarios, where extensive hyperparameter tuning is often impractical.   
\end{abstract}

\section{Introduction}
Graph Neural Networks (GNNs) \cite{Kipf2016SemiSupervisedCW,Wu2019ActiveLF} have emerged as a powerful tool for learning from graph-structured data, effectively capturing complex relationships and interdependencies between nodes. This progress has largely impacted areas where data inherently take the form of graphs, including social networks, biological networks, and communication systems. Concurrently, active learning \cite{Settles2008AnAO}, a subset of machine learning, has gained traction for its ability to efficiently utilize limited labeled data. In scenarios where labeling data is expensive or time-consuming, active learning strategically selects the most informative samples for labeling. This approach aims to maximize model prediction with a minimal amount of labeled data. The integration of active learning with GNNs presents a promising opportunity to enhance learning efficiency in graph-based learning tasks.
Recent approaches in active learning, particularly in the context of graph-structured data \cite{Gu2012TowardsAL,Ji2012AVM,Wu2019ActiveLF}, have focused on various strategies to identify the most informative nodes. These methods often revolve around uncertainty sampling, diversity sampling, or a combination of both, aiming to select nodes that are either uncertain under the current model or are representative of the underlying data distribution. This integration has shown significant potential in improving the efficiency of GNNs, especially in semi-supervised learning settings where labeled data are scarce. However, these methods primarily leverage node features or embeddings, often overlooking the rich structural information inherent in graphs. 
Thus, despite the advancements, there remains a notable gap in research concerning the optimal exploitation of graph topology in active learning for GNNs. 
This gap highlights the need for novel active learning strategies that can harness both the feature and structural information in graph-structured data. 
To address this research gap, we propose a novel method that integrates community detection with active learning in GNNs. The proposed method employs the SCAN algorithm \cite{SCAN2007}, recognized for effective community detection, alongside the PageRank algorithm \cite{Page1999ThePC} for node selection. By focusing on the community structures identified by SCAN and the node relevance ascertained by PageRank, we aim to select nodes that are informative in terms of features and pivotal in the graph’s structure. In essence, the synergy of the SCAN algorithm and PageRank enables the selection of samples that are meaningful. Specifically, SCAN identifies local structures, whereas PageRank sheds light on the broader, global structures of graph data.
Through extensive experiments, we demonstrate that the proposed method outperforms existing active learning methods for GNNs for a wide range of annotation budgets. Furthermore, the experiment on computational complexity indicates that the proposed approach leads to a reduction in query time. Thus, we summarize the contributions of the study as follows:
\begin{itemize}
    \item We propose a novel active learning method for Graph Neural Networks (GNNs) that integrates the SCAN method\cite{SCAN2007} for community detection with the PageRank algorithm \cite{Page1999ThePC}. 
    \item Through extensive experiments, we demonstrate that the proposed method substantially outperforms existing active learning methods in GNNs across various annotation budgets. 
    \item Additionally, the proposed method shows a notable reduction in computational complexity compared to recent active learning approaches for GNNs \cite{Ma2022PartitionBasedAL}, which is crucial in real active learning implementation where the waiting time during the annotation process is one of the important factors.  
\end{itemize}
\section{Related Work}\label{related work}
Active learning is a field in machine learning that focuses on reducing the cost of annotation while keeping the model performance stable and it has been comprehensively studied by \cite{Settles2008AnAO}. In this section, we focus on active learning for graph-structured data. 
Early works in active learning for graph-structured data primarily focused on leveraging graph topology for selecting informative samples \cite{Bilgic2010ActiveLF,Gu2012TowardsAL,Ji2012AVM}. These methods typically relied on measures like node centrality, degree, and cluster-based sampling, under the assumption that nodes with higher centrality or those bridging clusters are more informative. For example, AGE \cite{Cai2017ActiveLF} evaluates the informativeness of nodes by linearly combining centrality, density, and uncertainty. Furthermore, ARNMAB extends this approach by dynamically learning the weights with a multi-armed bandit mechanism and maximizing the surrogate reward \cite{Gao2018ActiveDN}.
The other type of approach in active learning for graph neural networks is implementing partition or clustering as part of community detection. For example, FeatProp \cite{Wu2019ActiveLF} combines node feature propagation with K-Medoids clustering for sample selection. The study was supported by a theoretical bound analysis showing an improvement in performance over other methods. Recently, Ma et al \cite{Ma2022PartitionBasedAL}. introduced the partition-based methods GraphPart and GraphPartFar, which align with active learning algorithms in GNNs by focusing on selecting nodes for optimal coverage of feature or representation spaces, typically through clustering algorithms.
On the other hand, The proposed method improves the conventional community detection approach by specifically employing clustering as a community detection algorithm. The proposed method works by capturing the local structures of nodes through community detection, while the PageRank scoring assesses their global significance. 
\section{Problem Formulation}
\subsection{Node Classification on Attributed Graphs}
Graph theory offers a robust framework for modeling complex systems through structures known as attributed graphs. Specifically, an attributed graph is denoted as $G = (V, E, X)$, where $V$ represents the set of nodes, $E \subseteq V \times V$ denotes the set of edges, and $X$ is the set of node attributes. Each node $v \in V$ is associated with an attribute vector $\mathbf{x}_v \in \mathbb{R}^d$. The adjacency matrix $A \in \{0, 1\}^{n \times n}$, where $n = |V|$, encodes the connectivity between nodes, with $A_{ij} = 1$ if there is an edge between nodes $i$ and $j$, and $A_{ij} = 0$ otherwise.

Node classification in attributed graphs aims to assign labels to nodes based on their attributes and structural positions in the graph. This involves learning a function $f: V \rightarrow L$, where $L$ is the set of possible labels. The challenge is to effectively leverage the information encoded in the graph structure and node attributes for accurate classification.

\subsection{Graph Neural Networks (GNNs)}
Graph Neural Networks (GNNs) are a class of neural networks designed for processing graph-structured data. They operate by aggregating information from a node's neighbors to update its representation. Formally, a GNN learns a function $f(G, X) \rightarrow Y$, where $Y$ is the output matrix representing node-level predictions.

The learning process in a GNN involves updating node representations through successive layers. Let $H^{(k)}$ be the matrix of node representations at the $k$-th layer, with $H^{(0)} = X$. The update rule at each layer is given by:
\begin{equation}
H^{(k+1)} = \sigma(\tilde{A} H^{(k)} W^{(k)})
\end{equation}
where $\tilde{A}$ is the normalized adjacency matrix, $W^{(k)}$ is the weight matrix for layer $k$, and $\sigma$ is a non-linear activation function.

The objective in training a GNN for node classification is often to minimize a loss function, typically the cross-entropy loss for the classification problem, defined as:
\begin{equation}
\mathcal{L} = -\sum_{v \in V_L} \sum_{l \in L} y_{vl} \log \hat{y}_{vl}
\end{equation}
where $V_L \subseteq V$ is the set of labeled nodes, $y_{vl}$ is the true label of node $v$ for label $l$, and $\hat{y}_{vl}$ is the predicted probability of node $v$ being in class $l$.

\subsection{Active Learning Task for Graph Neural Networks}
In the active learning scenario for GNNs, the objective is to select a subset of nodes $ V_S \subseteq V $ to label that maximizes the performance of the GNN. The selection process is guided by an acquisition function $ \mathcal{A}: V \rightarrow \mathbb{R} $, which scores each unlabeled node based on its expected utility for improving the model. The challenge is to design $\mathcal{A}$ to account for both the graph structure and node features. The active learning process iteratively selects nodes, updates the model, and re-evaluates the remaining unlabeled nodes.
In this study, we incorporate community detection into the active learning framework. We define a community structure $C$ within the graph, and the acquisition function $ \mathcal{A}$ is designed to preferentially select nodes that are central or informative within their communities, based on the hypothesis that such nodes provide more valuable information for the GNN model.
\section{Proposed Method}
The proposed method in this study consists of two main parts: partitioning the graph into communities and selecting representative nodes from these communities based on their PageRank scores.
\subsection{Community Detection using the SCAN Algorithm}
The initial phase of community detection in graphs involves partitioning the network into distinct communities. This task is accomplished using the SCAN algorithm, a method recognized for its capability to identify densely connected subgraphs or communities in a network. Unlike modularity-based approaches, the SCAN algorithm relies on structural similarity and a shared neighbor approach for community detection.
\subsubsection{Structural Similarity Measure.}
The core of the SCAN algorithm is the structural similarity measure between nodes, defined as follows:
\begin{equation}
    S(i, j) = \frac{|N(i) \cap N(j)|}{\sqrt{|N(i)| \cdot |N(j)|}}
\end{equation}
In this equation, $N(i)$ and $N(j)$ represent the neighbor sets of nodes $i$ and $j $, respectively. The measure $S(i, j)$ quantifies the similarity based on the shared neighbors of the two nodes, normalized by the geometric mean of their degrees.
\subsubsection{Community Detection Criteria.}
The SCAN algorithm employs two parameters, $\epsilon$ and $\mu$, to determine community membership. A node $ i $ is in the same community as node $j$ if the following conditions are met:
\begin{equation}
    S(i, j) \geq \epsilon \quad \text{and} \quad |N(i) \cap N(j)| \geq \mu
\end{equation}
where $ \epsilon $ is a similarity threshold and $\mu$ is the minimum number of shared neighbors required for community formation. These parameters allow the SCAN algorithm to classify nodes into clusters, hubs, or outliers, based on their structural roles within the network.
\subsection{Node Selection Based on PageRank}
Upon successfully partitioning the graph into communities, the next step is to select representative nodes from each community. This selection is based on the PageRank algorithm, which assigns a numerical weighting to each node in the network. The weight of a node is indicative of the probability of arriving at that node by randomly walking through the network. The PageRank \( PR(u) \) of a node \( u \) is defined as:
\begin{equation}
   PR(u) = \frac{1 - d}{N} + d \sum_{v \in B(u)} \frac{PR(v)}{L(v)} 
\end{equation}
where \( d \) is the damping factor, \( N \) is the total number of nodes, \( B(u) \) is the set of nodes that link to \( u \), and \( L(v) \) is the number of links from node \( v \). In the implementation, we use a damping factor of 0.95. For each community detected by the SCAN algorithm, the node with the highest PageRank score is selected. If the number of communities is less than the labeling budget \( b \), additional nodes with the next highest PageRank scores are selected until \( b \) nodes are chosen.
The final output is a set of \( b \) nodes, each representing its respective community, selected based on their significance within the network as determined by the PageRank algorithm. 
\subsection{SPA Algorithm}
The proposed \texttt{SPA} algorithm combines the strengths of the SCAN and PageRank algorithms. Algorithm \ref{alg:SPA} illustrates the detailed process of the proposed method SPA.
\begin{algorithm}
\caption{SPA Algorithm}
\begin{algorithmic}[1]
    \REQUIRE Adjacency matrix of the graph $A$, damping factor $\alpha$ for PageRank,\\ labeling budget $b$, clustering threshold $\epsilon$ and minimum neighbors $\mu$
    \ENSURE Sample of nodes $S$ to be labeled
    \STATE Initialize $S = \emptyset$
    \STATE Convert adjacency matrix $A$ to graph representation $G$
    \COMMENT{Community detection}
    \STATE Apply the SCAN algorithm to $G$ to detect communities $\{C_1, C_2, \ldots, C_k\}$ 
    \FOR{each community $C_i$ in $\{C_1, C_2, \ldots, C_k\}$}
        \STATE Calculate PageRank scores for all nodes in $C_i$ with damping factor $\alpha$ 
        \STATE Find node $n_{\text{max}}$ in $C_i$ with the highest PageRank score
        \STATE Add $n_{\text{max}}$ to $S$
    \ENDFOR
    \IF{$|S| < b$}
        \STATE Calculate PageRank for all nodes in $G \setminus S$
        \STATE Sort nodes in $G \setminus S$ by PageRank in descending order
        \STATE Add nodes from sorted list to $S$ until $|S| = b$
    \ENDIF
    \RETURN $S$ to be labeled
\end{algorithmic}\label{alg:SPA}
\end{algorithm}
\section{Experiments}\label{Experiments}
\subsection{Experiment Settings}
Experiments were conducted using various GNN models, including a 2-layer Graph Convolutional Network (GCN) \cite{Kipf2016SemiSupervisedCW} and a 2-layer GraphSAGE \cite{Hamilton2017InductiveRL}, both equipped with 16 hidden neurons. These models were trained using the Adam optimizer, starting with a learning rate of $1 \times 10^{-2}$ and a weight decay of $5 \times 10^{-4}$. In this study, we adopted a straightforward batch-active learning framework, labeling each sample within a batch, which corresponds to the defined budget in each experiment. The budget varies from 10 to 160 for smaller datasets and from 80 to 1280 for larger ones. Experiment results are presented as the mean of 10 independent runs, each with different random seeds. For transparency and reproducibility, the code is available on GitHub\footnote{\url{https://github.com/rickymaulanafajri/SPA}}.
\subsection{Dataset}
We conduct experiments on two standard node classification datasets: Citeseer and Pubmed \cite{Sen2008CollectiveCI}. Additionally, we use Corafull \cite{bojchevski2018deep} and WikiCS \cite{Mernyei2020WikiCSAW} to add diversity of the experiments. Furthermore, to assess the proposed method's performance on more heterophilous graphs, we experiment with the HeterophilousGraphDataset, which includes Minesweeper and Tolokers \cite{platonov2023critical}. Table \ref{tab:datasets} presents a summary and the statistics of the datasets used in this research. We use the \#Partitions parameter to determine the optimal number of communities in a graph, following the guidelines set by Ma et al. \cite{Ma2022PartitionBasedAL}.
\begin{table}[htbp]
\centering
\caption{Summary of datasets}
\begin{tabular}{c|c|c|c|c|c}
\hline \hline
Dataset    & \#Nodes & \#Edges & \#Features & \#Classes & \#Partitions \\ \hline
Citeseer   & 3,327   & 4,552   & 3,703      & 6         & 14           \\ \hline
Pubmed     & 19,717  & 44,324  & 500        & 3         & 8            \\ \hline
Corafull   & 19,793  & 126,842 & 8,710      & 70        & 7            \\ \hline
WikiCS   & 19,793  & 126,842 & 8,710      & 70        & 7            \\ \hline
Minesweeper   & 10,000  & 39,402 & 10      & 2        & 8            \\ \hline
Tolokers   & 11,758  & 519,000 & 10      & 2        & 7            \\ \hline

\hline
\end{tabular}\label{tab:datasets}
\end{table}
\subsection{Evaluation Metrics}
We use fundamental metrics such as accuracy and Macro-F1 score for evaluating the proposed method. Accuracy is measured as the ratio of correctly predicted class to and is mathematically represented as:
\begin{equation}
    \text{Accuracy} = \frac{TP + TN}{TP + TN + FP + FN}
\end{equation}
where $TP$, $TN$, $FP$, and $FN$ correspond to true positives, true negatives, false positives, and false negatives, respectively.
On the other hand, the Macro-F1 score is computed by taking the arithmetic mean of the F1 scores for each class independently. It is defined as:
\begin{equation}
    \text{Macro-F1} = \frac{1}{N} \sum_{i=1}^{N} \frac{2 \cdot \text{Precision}_i \cdot \text{Recall}_i}{\text{Precision}_i + \text{Recall}_i}
\end{equation}
where $N$ is the number of classes, and $ \text{Precision}_i$ and  $\text{Recall}_i$ denote the precision and recall for each class $i$, respectively. The precision and recall for each class are defined as follows:
\begin{equation}
    \text{Precision}_i = \frac{TP_i}{TP_i + FP_i}
\end{equation}
\begin{equation}
    \text{Recall}_i = \frac{TP_i}{TP_i + FN_i}
\end{equation}
\subsection{Baselines methods}
We compare the proposed approach with various active learning methods, divided into two types. 1. A general active learning method which works regardless of the graph neural network architecture such as Random, Uncertainty, and PageRank. 2. An active learning method that is specifically designed for graph-structured data such as Featprop, Graphpart, and GraphPartFar. The query strategy of each method is defined as follows:
\begin{itemize}
    \item Random: An active learning query strategy that selects a sample at random.
    \item Uncertainty \cite{Settles2008AnAO}: Select the nodes with the highest uncertainty based on the prediction model.
    \item PageRank \cite{Cai2017ActiveLF}: query the sample from a subset of points with the highest PageRank centrality score.
    \item FeatProp \cite{Wu2019ActiveLF}: First use KMeans to cluster the aggregated node features and then query the node closest to the center of the clustering.
    \item GraphPart and GraphPartFar \cite{Ma2022PartitionBasedAL}: A recent state-of-the-art method on active learning for graph structured data. GraphPart divides the graph into several partitions using Clauset-Newman-Moore greedy modularity maximization \cite{ClausetNewmanMoore2004} and selects the most representative sample in each partition to query. GraphPartFar increases the diversity of samples by selecting nodes that are not too close and similar to each other. 
\end{itemize}
\section{Results}\label{results}
\subsection{Experiment results of SPA on GCN}
\vspace{-10mm}
\begin{figure*}[!htb]
\centering
\subfloat[Citeseer ]{\includegraphics[width=.32\textwidth]{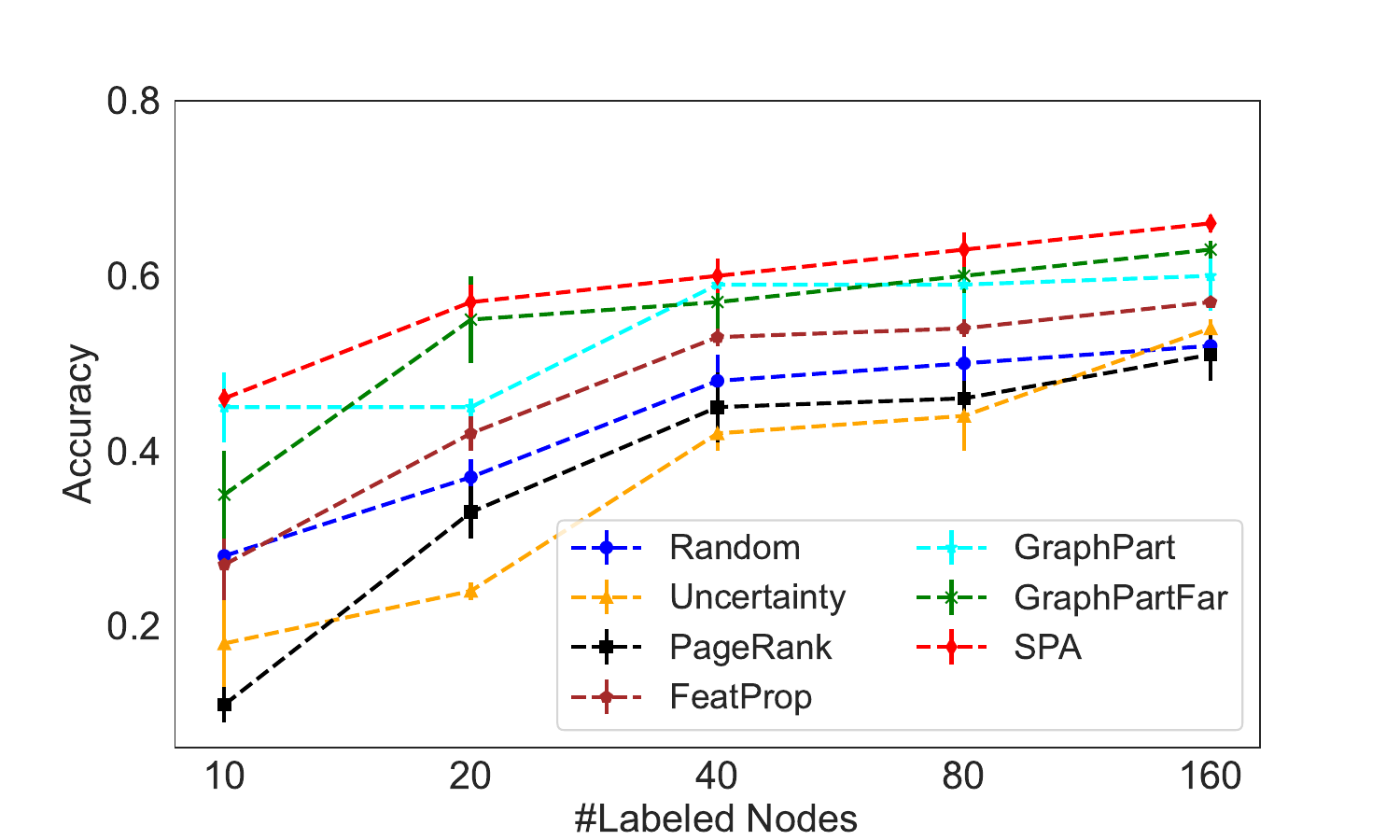}} \hspace{0.01\textwidth}
\subfloat[Pubmed]{\includegraphics[width = .32\textwidth]{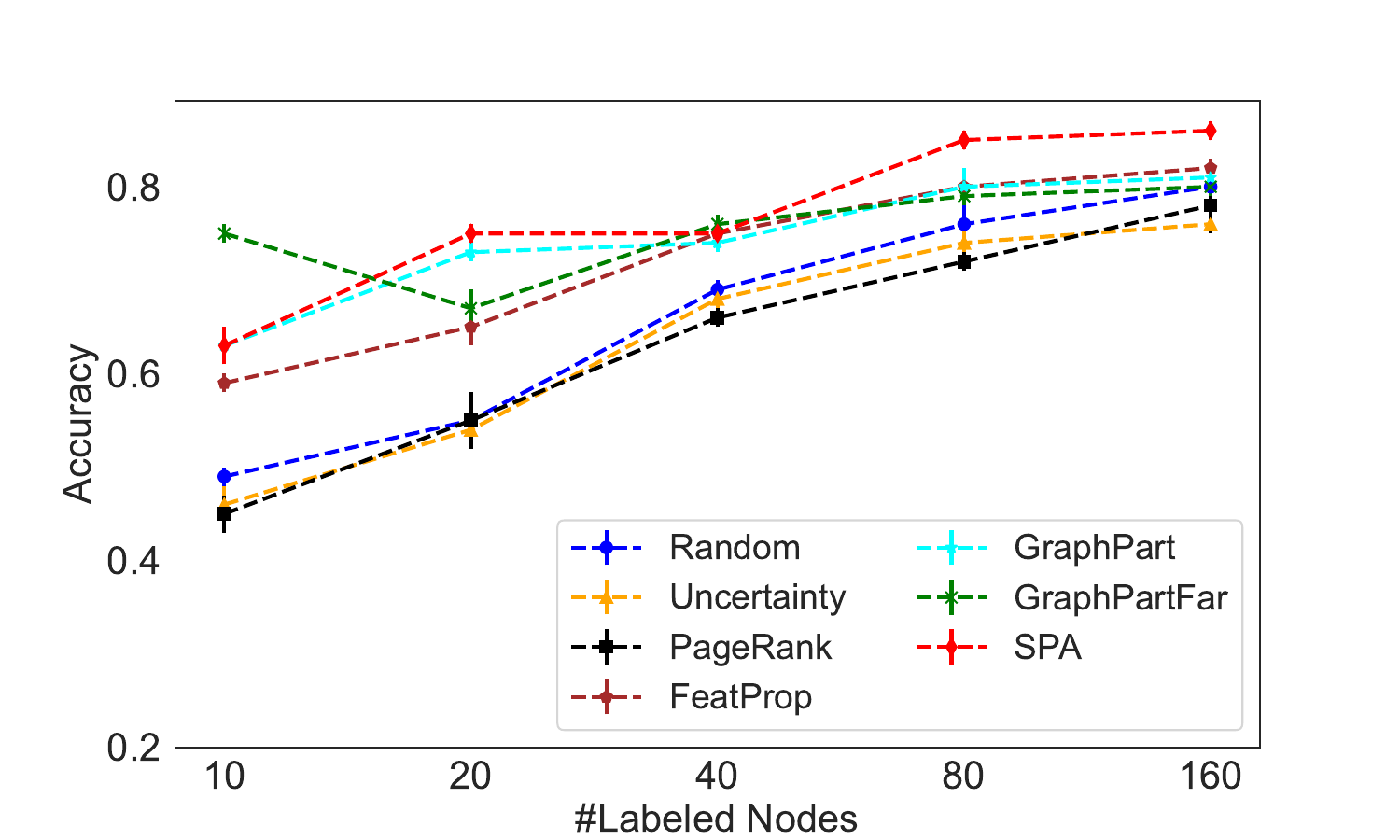}}\hspace{0.01\textwidth} 
\subfloat[Corafull]{\includegraphics[width =.32\textwidth]{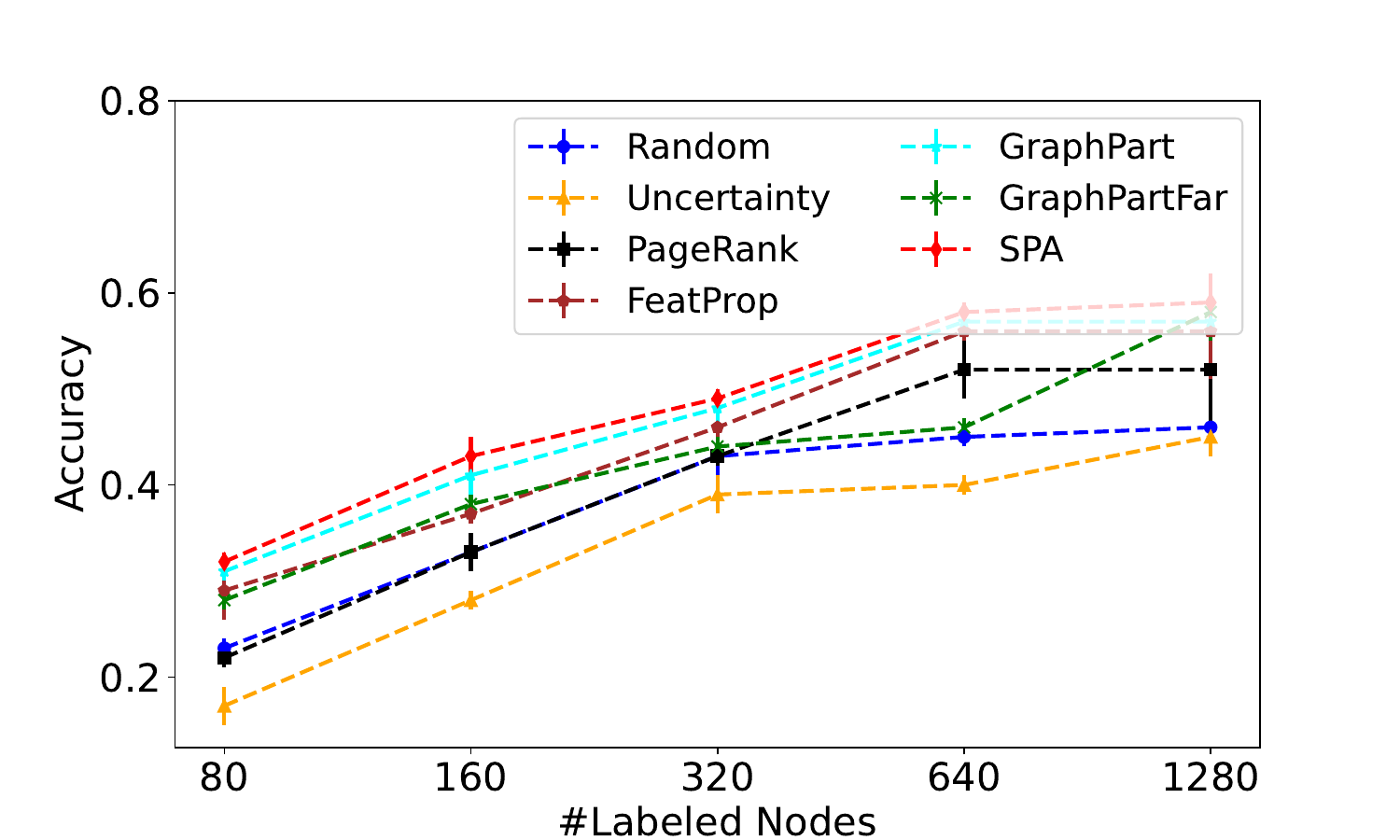}} \\
\subfloat[WikiCS]{\includegraphics[width = .32\textwidth]{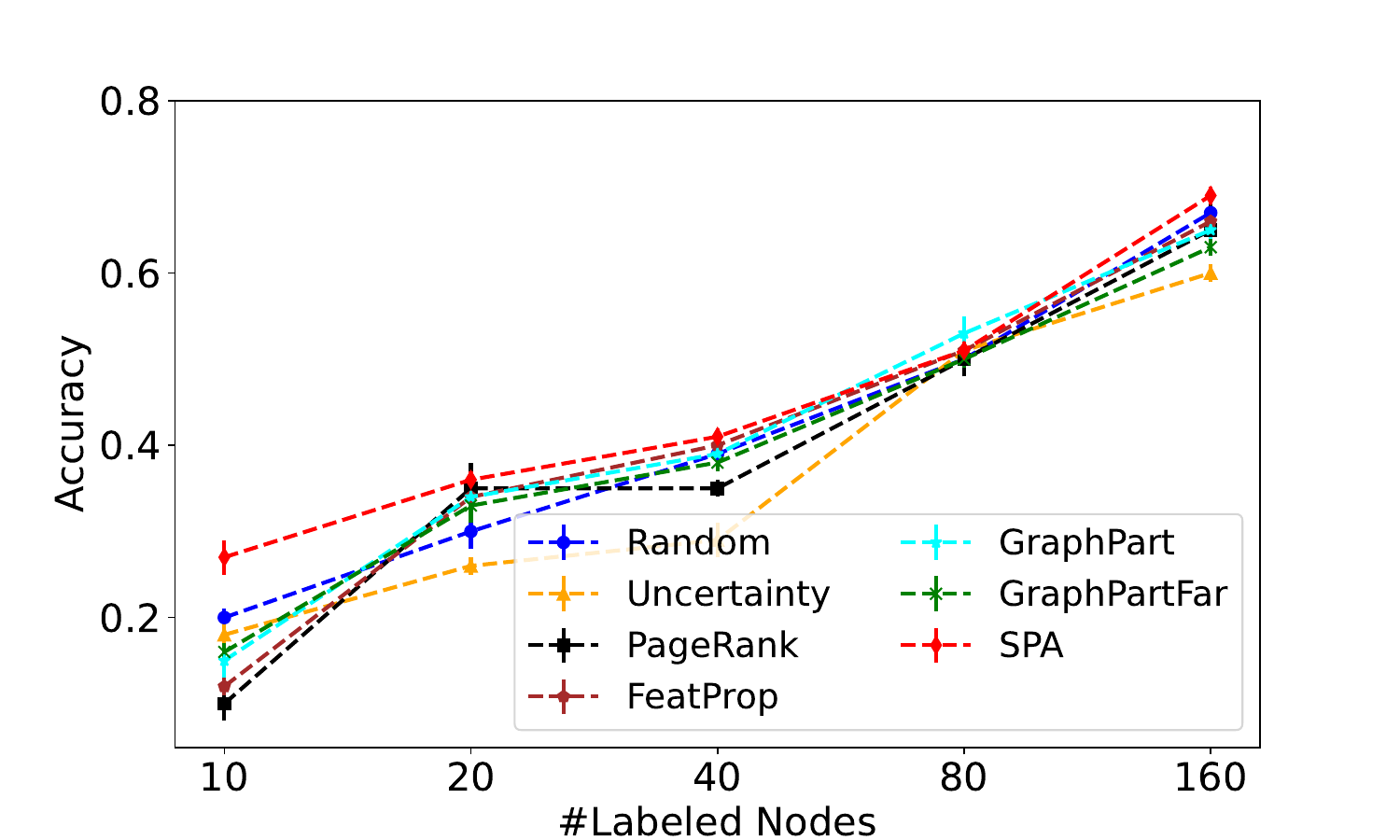}} \hspace{0.01\textwidth}
\subfloat[Minesweeper]{\includegraphics[width = .32\textwidth]{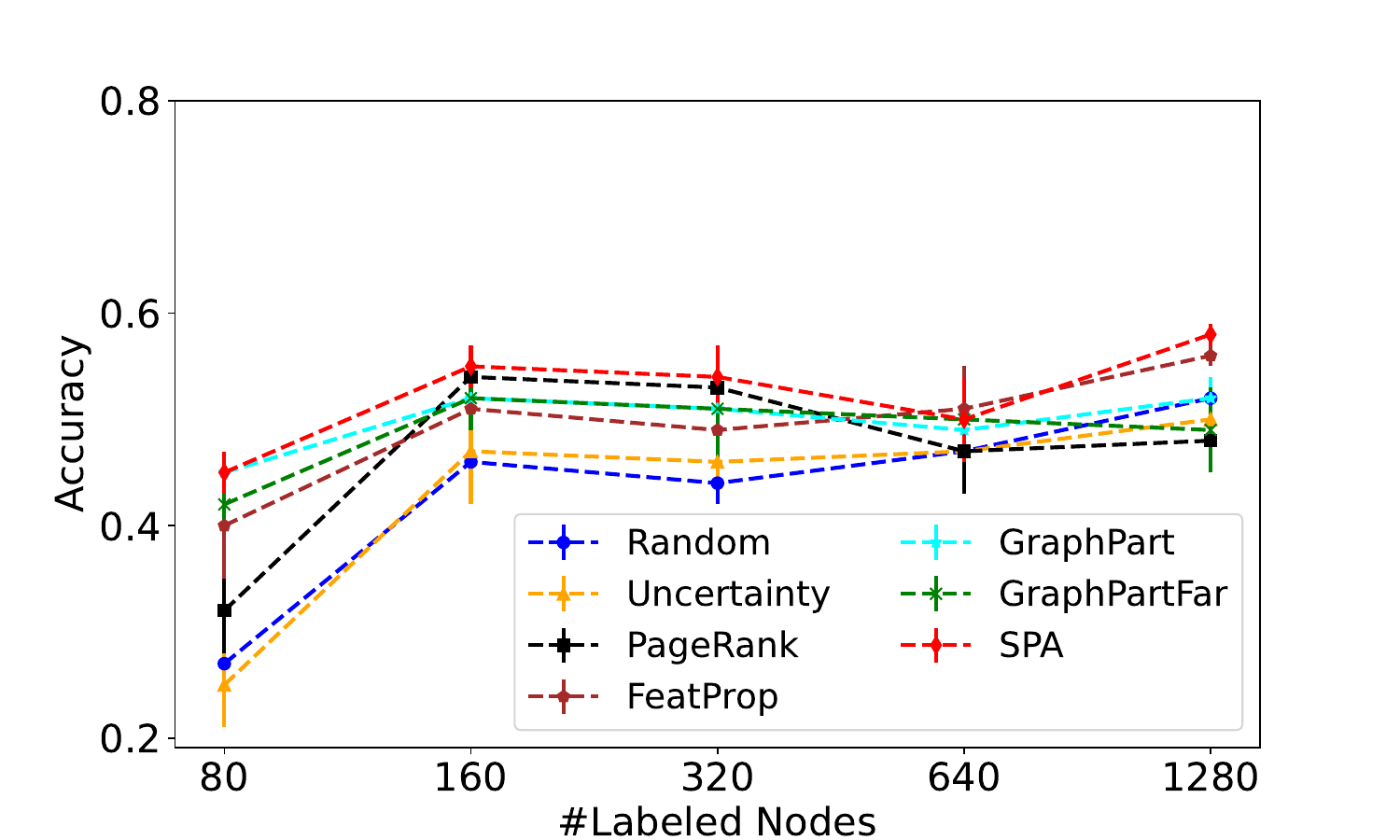}}\hspace{0.01\textwidth}
\subfloat[Tolokers]{\includegraphics[width = .32\textwidth]{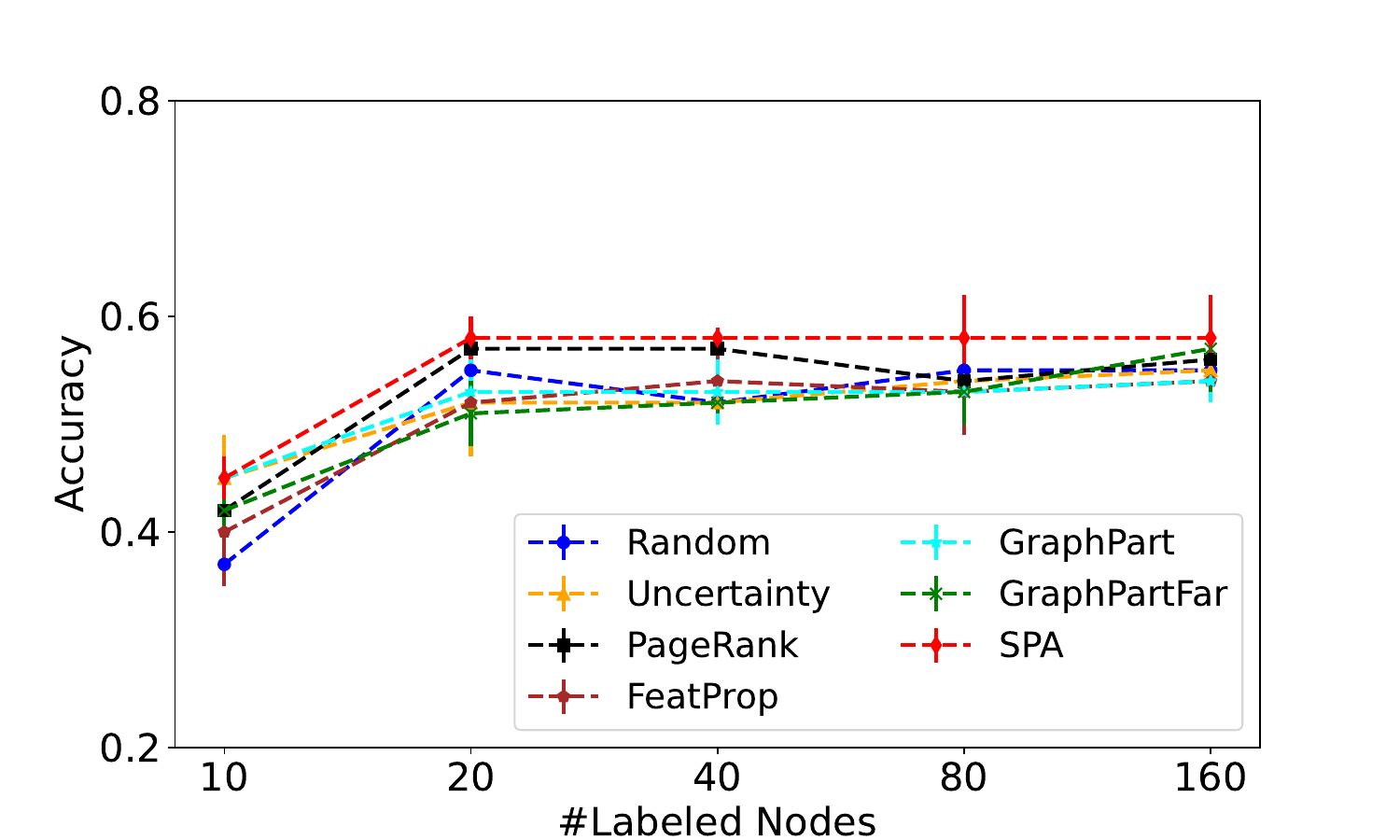}} \\
\caption{Comparative Evaluation of SPA and Baseline Methods Across Multiple Datasets: Accuracy Versus Number of Labeled Nodes in GCN Architecture.}\label{fig:Performance Comparison}
\end{figure*}

In this study, we present the active learning outcomes of the Graph Convolutional Network (GCN) across various datasets. First, we analyze the accuracy score of the proposed method compared to the baseline. Figure \ref{fig:Performance Comparison} illustrates the comparative evaluation of the proposed and baseline methods. It is evident that while all methods exhibit a general trend of improved accuracy with an increased budget, the proposed approach consistently maintains a higher accuracy rate. For example, in  Citeseer, Pubmed, and Corafull, the proposed method excels in accuracy from the smallest labeling budget to the largest one. Although, in the other dataset the accuracy of the proposed method only shows a marginal improvement, it is still higher compared to all the baselines. 
\begin{table}[tbp!]
\centering
\caption{Summary of the Macro-F1 score of the proposed approach using GCN architecture. The numerical values indicate the mean Macro-F1 score derived from 10 separate trials. The best score is in bold marker. }
\scalebox{0.65}{
\begin{tabular}{c|ccc|ccc|ccc}
\hline \hline
\multirow{2}{*}{Baselines} & \multicolumn{3}{c|}{Citeseer}                                                                                            & \multicolumn{3}{c|}{Pubmed}                                                                                              & \multicolumn{3}{c}{Corafull}                                                                                           \\ \cline{2-10} 
                           & \multicolumn{1}{c|}{20}                       & \multicolumn{1}{c|}{40}                       & 80                       & \multicolumn{1}{c|}{10}                       & \multicolumn{1}{c|}{20}                       & 40                       & \multicolumn{1}{c|}{10}                       & \multicolumn{1}{c|}{20}                        & 40                     \\ \hline \hline
Random                     & \multicolumn{1}{c|}{28.4 $\pm$12.6}            & \multicolumn{1}{c|}{37.6$\pm$6.7}            & 48.9$\pm$5.8            & \multicolumn{1}{c|}{49.1$\pm$11.4}           & \multicolumn{1}{c|}{55.7$\pm$10.6}           & 69.5$\pm$6.2            & \multicolumn{1}{c|}{23.8$\pm$2.0}            & \multicolumn{1}{c|}{33.2$\pm$1.8}             & 43.2$\pm$2.1          \\ \hline
Uncertainty                & \multicolumn{1}{c|}{18.8$\pm$7.1}            & \multicolumn{1}{c|}{24.2$\pm$5.7}            & 42.8$\pm$12.5           & \multicolumn{1}{c|}{46.0$\pm$11.5}           & \multicolumn{1}{c|}{54.7$\pm$10.4}           & 64.8$\pm$9.2            & \multicolumn{1}{c|}{17.3$\pm$1.7}            & \multicolumn{1}{c|}{28.1$\pm$2.1}             & 39.1$\pm$1.2          \\ \hline
PageRank                   & \multicolumn{1}{c|}{27.2$\pm$6.6}            & \multicolumn{1}{c|}{36.3$\pm$7.8}            & 49.6$\pm$7.5            & \multicolumn{1}{c|}{45.3$\pm$8.5}            & \multicolumn{1}{c|}{55.7$\pm$13.0}           & 66.3$\pm$8.6            & \multicolumn{1}{c|}{22.8$\pm$1.4}            & \multicolumn{1}{c|}{33.4$\pm$1.1}             & 43.7$\pm$0.5          \\ \hline
FeatProp                   & \multicolumn{1}{c|}{27.9$\pm$5.5}            & \multicolumn{1}{c|}{42.9$\pm$4.5}            & 53.7$\pm$4.5            & \multicolumn{1}{c|}{59.1$\pm$5.5}            & \multicolumn{1}{c|}{65.4$\pm$5.2}            & 75.1$\pm$2.8            & \multicolumn{1}{c|}{29.6$\pm$1.0}            & \multicolumn{1}{c|}{37.6$\pm$0.8}             & 46.7$\pm$0.8          \\ \hline
GraphPart                  & \multicolumn{1}{c|}{45.0$\pm$0.7}            & \multicolumn{1}{c|}{45.4$\pm$4.1}            & 59.0$\pm$2.0            & \multicolumn{1}{c|}{63.0$\pm$0.7}            & \multicolumn{1}{c|}{73.2$\pm$1.0}            & 74.0$\pm$1.3            & \multicolumn{1}{c|}{31.0$\pm$1.3}            & \multicolumn{1}{c|}{\textbf{41.2$\pm$1.4}}    & \textbf{48.6$\pm$0.5} \\ \hline
GraphPartFar               & \multicolumn{1}{c|}{35.1$\pm$0.6}            & \multicolumn{1}{c|}{55.2$\pm$2.4}            & 57.5$\pm$2.0            & \multicolumn{1}{c|}{75.7$\pm$0.3}            & \multicolumn{1}{c|}{67.5$\pm$0.5}            & 76.2$\pm$0.9            & \multicolumn{1}{c|}{28.0$\pm$1.2}            & \multicolumn{1}{c|}{38.4$\pm$0.6}             & 44.3$\pm$0.7          \\ \hline
SPA                        & \multicolumn{1}{c|}{\textbf{46.3$\pm$0.5}}   & \multicolumn{1}{c|}{\textbf{57.1$\pm$0.1}}   & \textbf{60.8$\pm$2.4}   & \multicolumn{1}{c|}{\textbf{63.1$\pm$0.1}}   & \multicolumn{1}{c|}{\textbf{75.1$\pm$0.4}}   & \textbf{77.5$\pm$3.2}   & \multicolumn{1}{c|}{\textbf{31.3$\pm$0.2}}   & \multicolumn{1}{c|}{40.4$\pm$0.2}             & 47.7$\pm$0.2          \\ \hline \hline
\multirow{2}{*}{Baselines} & \multicolumn{3}{c|}{WikiCS}                                                                                              & \multicolumn{3}{c|}{Minesweeper}                                                                                         & \multicolumn{3}{c}{Tolokers}                                                                                           \\ \cline{2-10} 
                           & \multicolumn{1}{c|}{20}                       & \multicolumn{1}{c|}{40}                       & 80                       & \multicolumn{1}{c|}{160}                      & \multicolumn{1}{c|}{320}                      & 640                      & \multicolumn{1}{c|}{20}                       & \multicolumn{1}{c|}{40}                        & 80                     \\ \hline
Random                     & \multicolumn{1}{c|}{30.2 $\pm$ 2.1}             & \multicolumn{1}{c|}{51.3 $\pm$ 4.5}             & 50.2$\pm$1.2            & \multicolumn{1}{c|}{46.7$\pm$6.7}            & \multicolumn{1}{c|}{44.4$\pm$6.5}            & 47.1$\pm$4.5            & \multicolumn{1}{c|}{55.2$\pm$8.9}            & \multicolumn{1}{c|}{52.4$\pm$0.66}           & 55.4$\pm$0.30        \\ \hline
Uncertainty                & \multicolumn{1}{c|}{26.4$\pm$1.2}            & \multicolumn{1}{c|}{29.3$\pm$0.7}            & 51.8$\pm$ 2.4           & \multicolumn{1}{c|}{47.1$\pm$0.55}          & \multicolumn{1}{c|}{46.0$\pm$1.41}          & 47.3$\pm$0.02          & \multicolumn{1}{c|}{52.4$\pm$0.77}          & \multicolumn{1}{c|}{52.8$\pm$0.93}           & 54.3$\pm$0.81        \\ \hline
PageRank                   & \multicolumn{1}{c|}{35.4$\pm$4.3}            & \multicolumn{1}{c|}{35.7$\pm$0.8}            & 50.4$\pm$5.4            & \multicolumn{1}{c|}{54.6$\pm$0.41}          & \multicolumn{1}{c|}{53.9$\pm$0.49}          & 47.3$\pm$0.07          & \multicolumn{1}{c|}{57.9$\pm$0.56}          & \multicolumn{1}{c|}{57.1$\pm$0.41}           & 54.4$\pm$0.99        \\ \hline
FeatProp                   & \multicolumn{1}{c|}{34.2$\pm$2.1}            & \multicolumn{1}{c|}{40.1$\pm$0.41}          & 51.4$\pm$0.06          & \multicolumn{1}{c|}{51.3$\pm$ 0.24}         & \multicolumn{1}{c|}{49.0$\pm$0.05}          & 51.3$\pm$0.02          & \multicolumn{1}{c|}{52.3$\pm$0.45}          & \multicolumn{1}{c|}{54.3$\pm$0.72}           & 53.3$\pm$0.45        \\ \hline
GraphPart                  & \multicolumn{1}{c|}{34.2$\pm$0.45}          & \multicolumn{1}{c|}{39.0$\pm$ 0.21}         & \textbf{53.2$\pm$0.44} & \multicolumn{1}{c|}{52.2$\pm$0.03}          & \multicolumn{1}{c|}{51.0$\pm$0.93}          & 49.8$\pm$0.84          & \multicolumn{1}{c|}{53.2$\pm$0.48}          & \multicolumn{1}{c|}{53.3$\pm$0.32}           & 53.4$\pm$0.21        \\ \hline
GraphPartFar               & \multicolumn{1}{c|}{33.4$\pm$0.05}          & \multicolumn{1}{c|}{38.0$\pm$0.1}           & 50.1$\pm$0.21          & \multicolumn{1}{c|}{52.3$\pm$6.0}           & \multicolumn{1}{c|}{51.7$\pm$0.16}          & 50.3$\pm$0.29          & \multicolumn{1}{c|}{51.7$\pm$ 0.24}         & \multicolumn{1}{c|}{52.2$\pm$0.18}           & 53.4$\pm$0.07        \\ \hline \hline
SPA                        & \multicolumn{1}{c|}{\textbf{36.5$\pm$0.78}} & \multicolumn{1}{c|}{\textbf{41.5$\pm$0.02}} & 51.2$\pm$0.45          & \multicolumn{1}{c|}{\textbf{55.1$\pm$0.07}} & \multicolumn{1}{c|}{\textbf{54.3$\pm$0.45}} & \textbf{52.2$\pm$0.45} & \multicolumn{1}{c|}{\textbf{58.0$\pm$0.06}}  & \multicolumn{1}{c|}{\textbf{58.0$\pm$ 0.04}} & \textbf{58.0 0.02}    \\ \hline \hline
\end{tabular}}\label{tab:resultsgcn}
\end{table}
Next, we examine the Macro-F1 efficacy of the proposed method. The results of these experiments are summarized in Table \ref{tab:resultsgcn}. Notably, the SPA method demonstrates consistently high results across various datasets, including Citeseer, Pubmed, and Corafull. Its superiority is particularly evident in situations with diverse sample sizes. This is most clearly observed in the Citeseer dataset, where SPA excels within a 40-label budget, highlighting its effectiveness in moderately sized sample environments. While GraphPart shows a competitive edge, particularly in the 20-label budget scenario of the Corafull dataset, SPA still maintains a slight but significant advantage in the 10-label budget.
\subsection{Experiment Result of SPA on GraphSAGE}
\begin{figure*}[!htb]
\centering
\subfloat[Citeseer ]{\includegraphics[width=.32\textwidth]{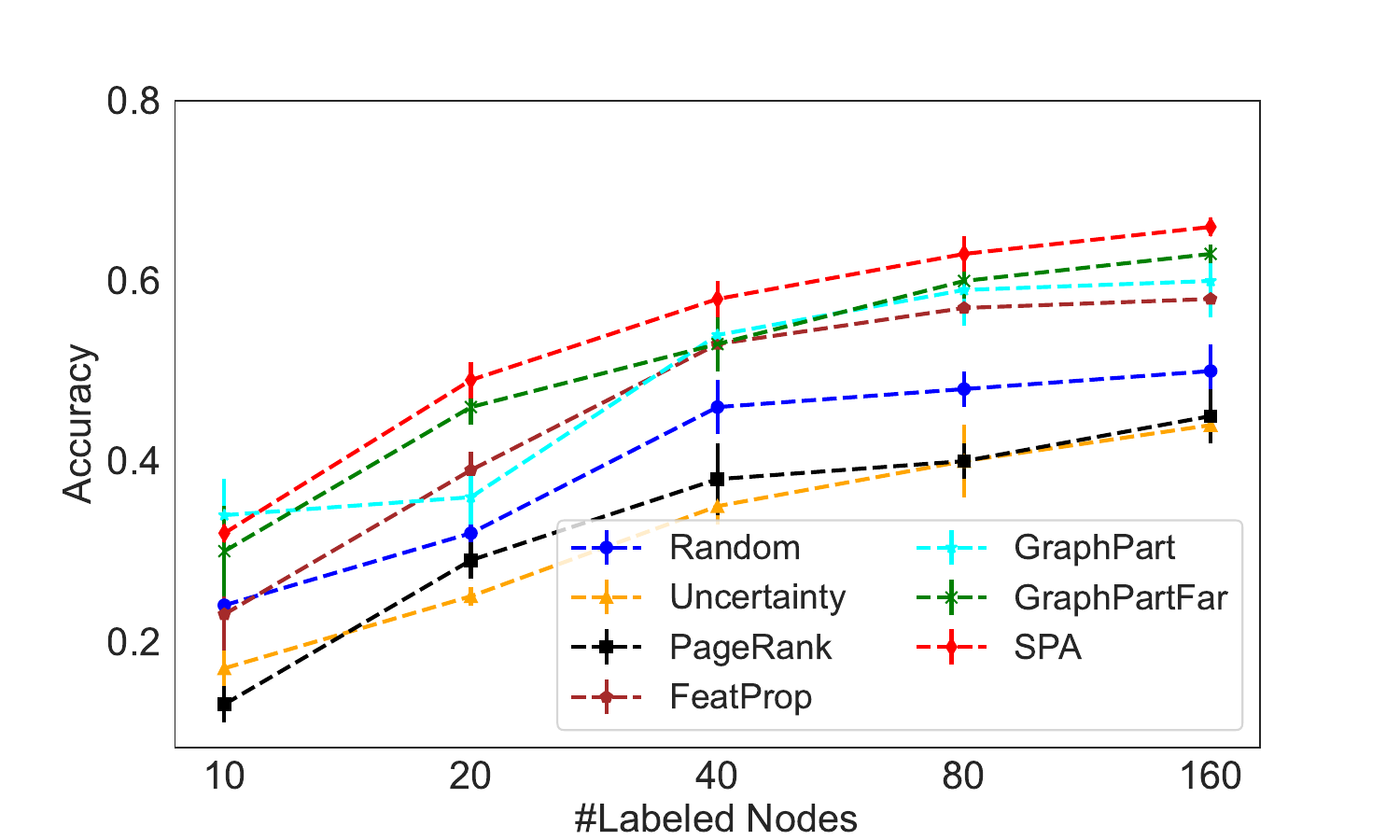}} \hspace{0.01\textwidth}
\subfloat[Pubmed]{\includegraphics[width = .32\textwidth]{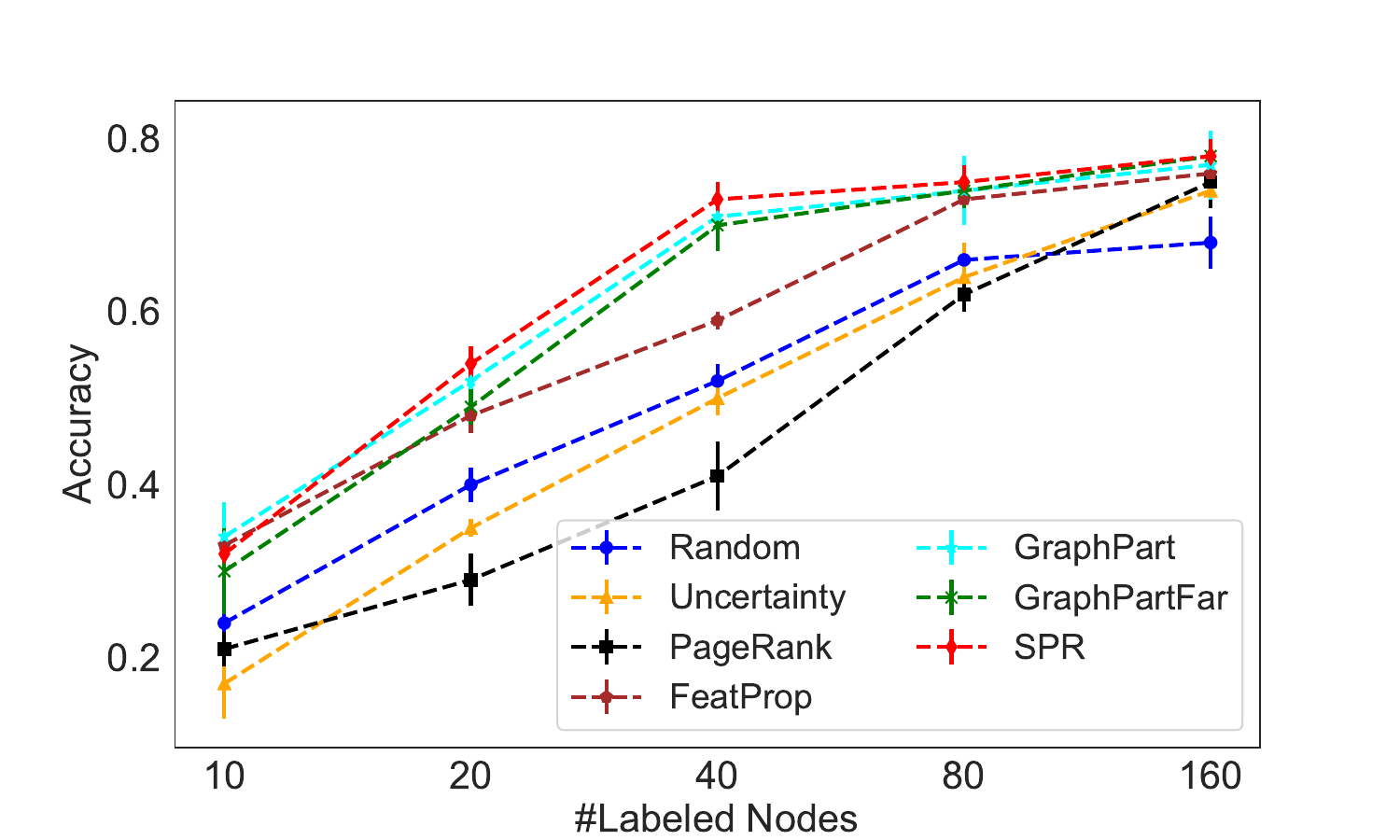}}\hspace{0.01\textwidth} 
\subfloat[Corafull]{\includegraphics[width =.32\textwidth]{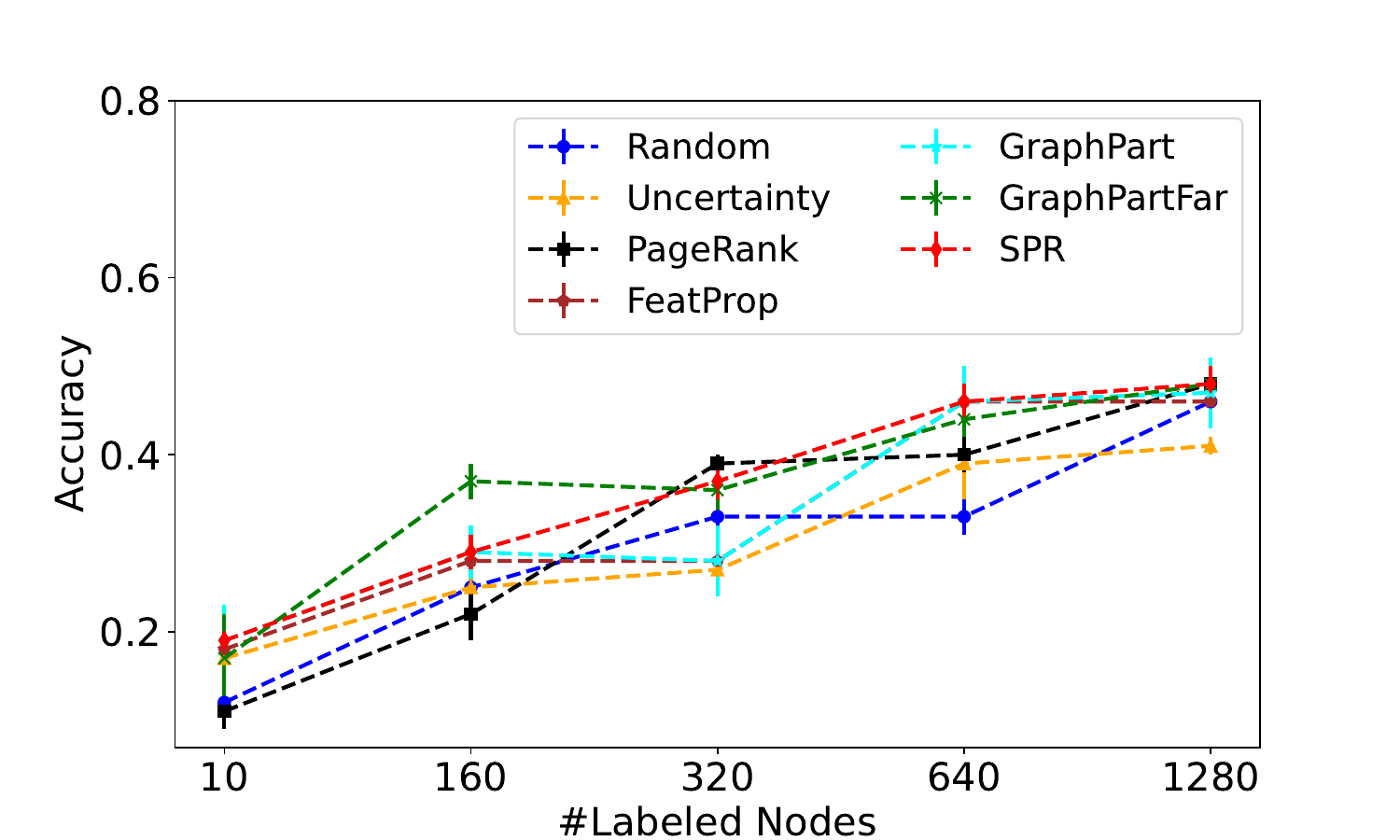}} \\
\subfloat[WikiCS]{\includegraphics[width = .32\textwidth]{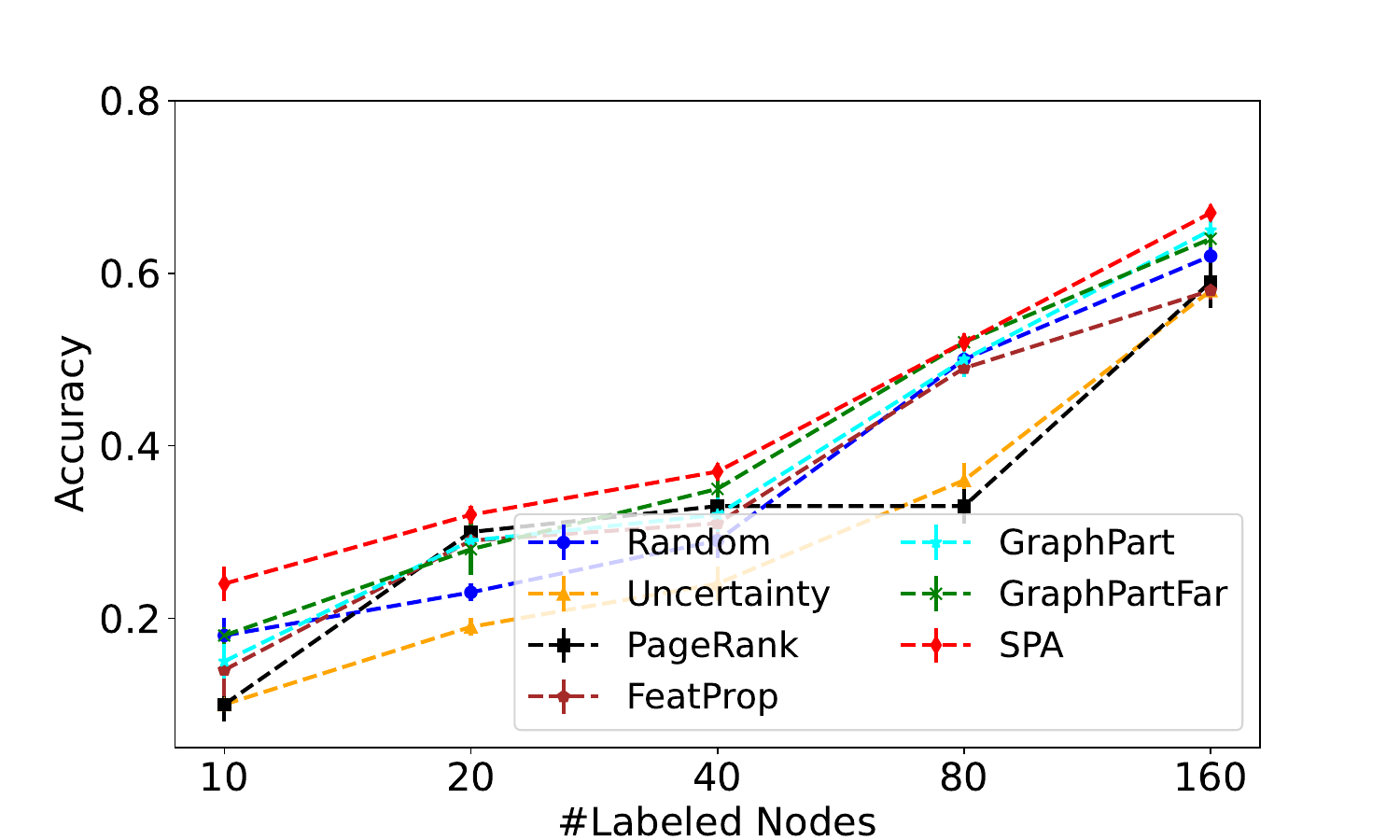}} \hspace{0.01\textwidth}
\subfloat[Minesweeper]{\includegraphics[width = .32\textwidth]{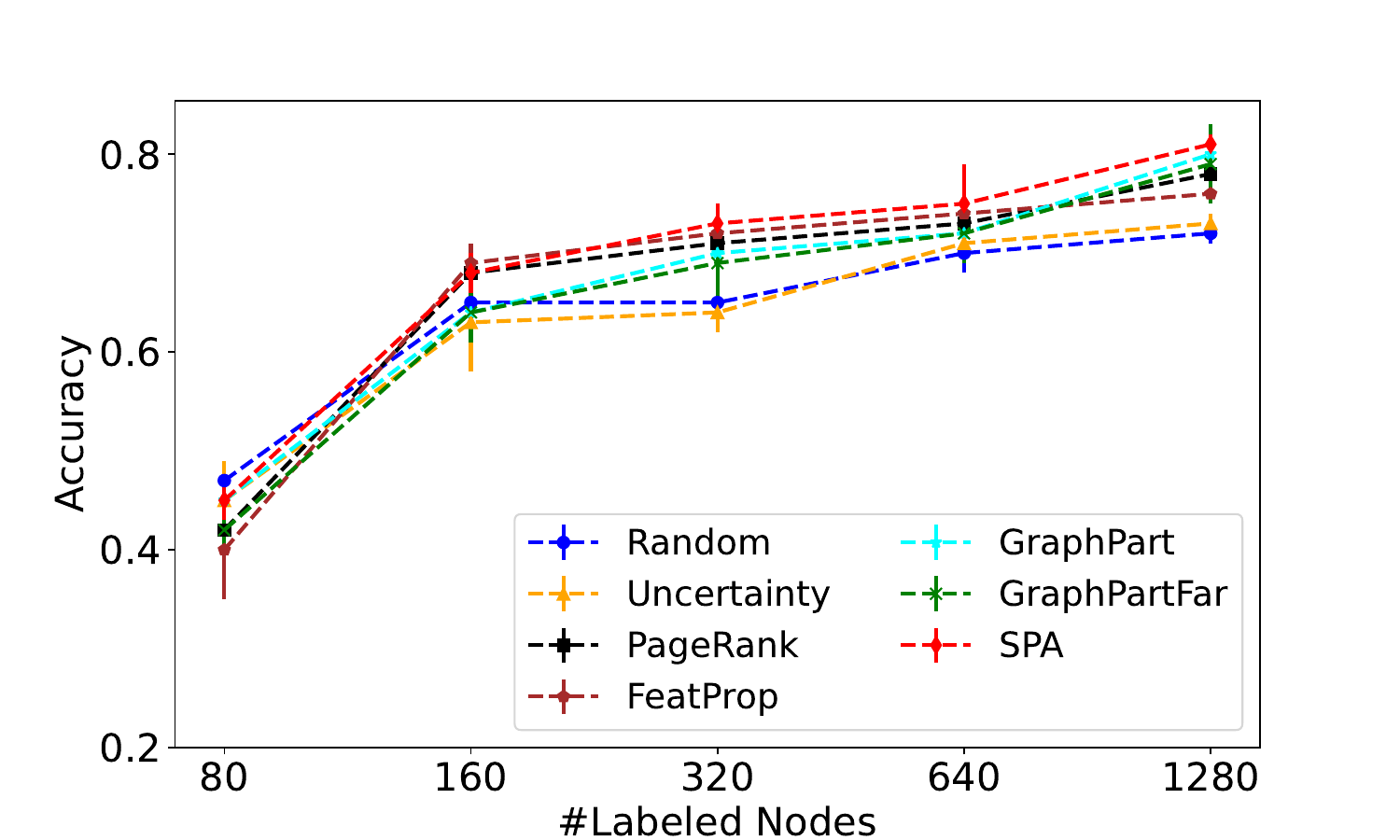}}\hspace{0.01\textwidth}
\subfloat[Tolokers]{\includegraphics[width = .32\textwidth]{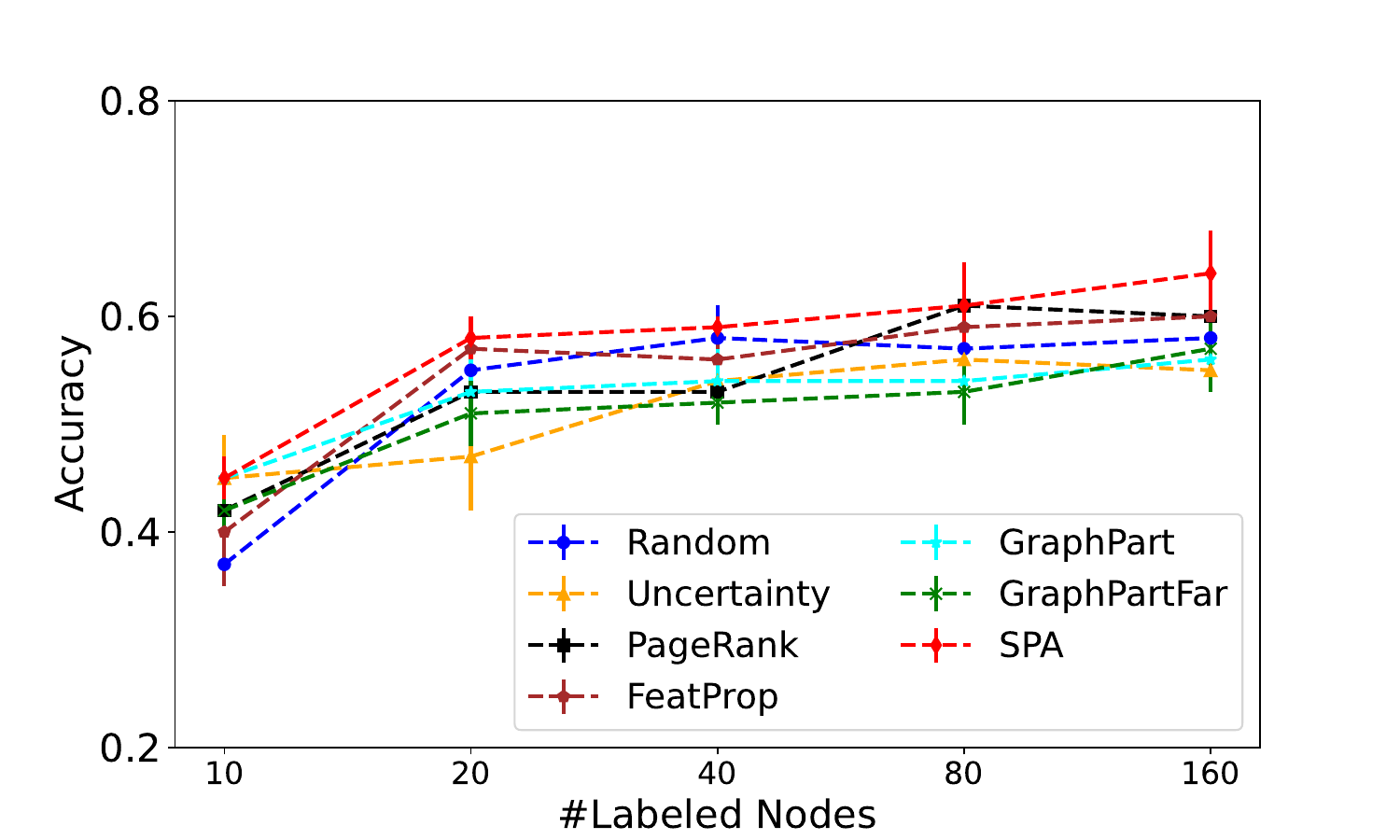}}  \\
\caption{Comparative Evaluation of SPA and Baseline Methods Across Multiple Datasets: Accuracy vs. Number of Labeled Nodes in GraphSAGE Architecture.}\label{fig:Performance Comparison sage}
\end{figure*}

In addition, we conducted further experimental analysis using another Graph Neural Network (GNN) architecture, specifically GraphSAGE. We use all the datasets from the previous experiment. The initial focus was on evaluating the accuracy score of the proposed methods within the GraphSAGE architecture. Figure \ref{fig:Performance Comparison sage} illustrates the effectiveness of the proposed method in achieving higher accuracy. For instance, in the Citeseer dataset, the proposed method begins with an accuracy score of 0.32 at a 10-label budget, and it reaches its peak accuracy score with a score of 0.68 at a 1280-label budget.
Secondly, we compare the Macro-F1 score of each method.  Table \ref{tab:resultssage} shows the Macro-F1 score of the baseline methods compared with the proposed approach. Table \ref{tab:resultssage} illustrates that the proposed method consistently outperformed existing models in terms of Macro-F1 score in GraphSAGE architecture. Notably, it demonstrated a significant improvement in Macro-F1 scores in WikiCS, Minesweeper, and Tolokers.
\vspace{-5mm}
\begin{table*}[htbp!]
\centering
\caption{Summary of the Macro-F1 score of the proposed approach using GraphSAGE architecture.}
\scalebox{0.65}{
\begin{tabular}{c|ccc|ccc|ccc}
\hline \hline
\multirow{2}{*}{Baselines} & \multicolumn{3}{c|}{Citeseer}                                                                                                     & \multicolumn{3}{c|}{Pubmed}                                                                                                        & \multicolumn{3}{c}{Corafull}                                                                                                     \\ \cline{2-10} 
                           & \multicolumn{1}{c|}{10}                          & \multicolumn{1}{c|}{20}                          & 40                          & \multicolumn{1}{c|}{20}                          & \multicolumn{1}{c|}{40}                           & 80                          & \multicolumn{1}{c|}{160}                         & \multicolumn{1}{c|}{320}                         & 640                         \\ \hline
Random                     & \multicolumn{1}{c|}{24.1 $\pm$ 11.9}         & \multicolumn{1}{c|}{32.4 $\pm$ 6.6}          & 46.1 $\pm$ 5.6          & \multicolumn{1}{c|}{40.6 $\pm$ 13.0}         & \multicolumn{1}{c|}{52.3 $\pm$ 12.2}          & 66.3 $\pm$ 7.9          & \multicolumn{1}{c|}{15.4 $\pm$ 1.1}          & \multicolumn{1}{c|}{23.4 $\pm$ 1.4}          & 33.5 $\pm$ 1.3          \\ \hline
Uncertainty                & \multicolumn{1}{c|}{17.6 $\pm$ 6.3}          & \multicolumn{1}{c|}{25.1 $\pm$ 6.1}          & 35.7 $\pm$ 4.4          & \multicolumn{1}{c|}{35.2 $\pm$ 6.6}          & \multicolumn{1}{c|}{50.5 $\pm$ 10.1}          & 64.1 $\pm$ 10.5         & \multicolumn{1}{c|}{15.3 $\pm$ 0.8}          & \multicolumn{1}{c|}{27.0 $\pm$ 1.9}          & 39.7 $\pm$ 1.1          \\ \hline
PageRank                   & \multicolumn{1}{c|}{13.0 $\pm$ 1.0}          & \multicolumn{1}{c|}{29.8 $\pm$ 1.3}          & 38.3 $\pm$ 2.2          & \multicolumn{1}{c|}{29.7 $\pm$ 0.3}          & \multicolumn{1}{c|}{41.7 $\pm$ 0.6}           & 62.9 $\pm$ 0.3          & \multicolumn{1}{c|}{12.4 $\pm$ 0.3}          & \multicolumn{1}{c|}{19.4 $\pm$ 0.8}          & 30.3 $\pm$ 0.4           \\ \hline
Featprop                   & \multicolumn{1}{c|}{23.4 $\pm$ 4.3}          & \multicolumn{1}{c|}{39.9 $\pm$ 6.2}          & 53.5 $\pm$ 3.3          & \multicolumn{1}{c|}{48.0 $\pm$ 5.9}          & \multicolumn{1}{c|}{59.1 $\pm$ 6.0}           & 73.6 $\pm$ 1.7          & \multicolumn{1}{c|}{18.7 $\pm$ 0.8}          & \multicolumn{1}{c|}{25.8 $\pm$ 0.7}          & 35.0 $\pm$ 1.6          \\ \hline
GraphPart                  & \multicolumn{1}{c|}{\textbf{34.1 $\pm$ 6.4}} & \multicolumn{1}{c|}{36.1 $\pm$ 6.4}          & 54.0 $\pm$ 4.6          & \multicolumn{1}{c|}{52.0 $\pm$ 0.8}          & \multicolumn{1}{c|}{71.5 $\pm$ 0.5}           & 74.6 $\pm$ 1.1          & \multicolumn{1}{c|}{\textbf{19.7 $\pm$ 0.9}} & \multicolumn{1}{c|}{\textbf{28.3 $\pm$ 0.7}} & 36.1 $\pm$ 1.0          \\ \hline
GraphPartFar               & \multicolumn{1}{c|}{30.7 $\pm$ 2.3}          & \multicolumn{1}{c|}{46.9 $\pm$ 5.0}          & 53.1 $\pm$ 4.0          & \multicolumn{1}{c|}{49.7 $\pm$ 3.1}          & \multicolumn{1}{c|}{70.7 $\pm$ 1.6}           & 74.2 $\pm$ 0.4          & \multicolumn{1}{c|}{17.5 $\pm$ 1.0}          & \multicolumn{1}{c|}{26.2 $\pm$ 1.4}          & 34.1 $\pm$ 0.9          \\ \hline \hline
SPA                        & \multicolumn{1}{c|}{32.6 $\pm$ 0.2}          & \multicolumn{1}{c|}{\textbf{49.5 $\pm$ 1.3}} & \textbf{58.2 $\pm$ 2.1} & \multicolumn{1}{c|}{\textbf{54.1 $\pm$ 2.1}} & \multicolumn{1}{c|}{\textbf{73.20 $\pm$ 2.4}} & \textbf{75.0 $\pm$ 0.3} & \multicolumn{1}{c|}{19.2 $\pm$ 0.2}          & \multicolumn{1}{c|}{27.9 $\pm$ 2.3}          & \textbf{36.7 $\pm$ 0.2} \\ \hline \hline
\multirow{2}{*}{Baselines} & \multicolumn{3}{c|}{WikiCS}                                                                                                       & \multicolumn{3}{c|}{Minesweeper}                                                                                                   & \multicolumn{3}{c}{Tolokers}                                                                                                     \\ \cline{2-10} 
                           & \multicolumn{1}{c|}{20}                          & \multicolumn{1}{c|}{40}                          & 80                          & \multicolumn{1}{c|}{160}                         & \multicolumn{1}{c|}{320}                          & 640                         & \multicolumn{1}{c|}{20}                          & \multicolumn{1}{c|}{40}                          & 80                          \\ \hline
Random                     & \multicolumn{1}{c|}{23.2 $\pm$ 2.3}          & \multicolumn{1}{c|}{29.03 $\pm$ 0.3}         & 50.3 $\pm$ 0.4          & \multicolumn{1}{c|}{65.2 $\pm$ 1.2}          & \multicolumn{1}{c|}{65.4 $\pm$ 0.3}           & 70.2 $\pm$ 0.2          & \multicolumn{1}{c|}{55.3 $\pm$ 0.6}          & \multicolumn{1}{c|}{58.2 $\pm$ 0.8}          & 57.4 $\pm$ 0.5          \\ \hline
Uncertainty                & \multicolumn{1}{c|}{19.2 $\pm$ 0.9}          & \multicolumn{1}{c|}{24.4 $\pm$ 0.2}          & 36.2 $\pm$ 0.8          & \multicolumn{1}{c|}{63.2 $\pm$ 0.7}          & \multicolumn{1}{c|}{64.6 $\pm$ 1.3}           & 71.4 $\pm$ 2.4          & \multicolumn{1}{c|}{47.1 $\pm$ 8.3}          & \multicolumn{1}{c|}{54.2 $\pm$ 5.2}          & 56.5 $\pm$ 6.5          \\ \hline
PageRank                   & \multicolumn{1}{c|}{30.2 $\pm$ 3.2}          & \multicolumn{1}{c|}{33.6 $\pm$ 2.4}          & 33.8 $\pm$ 7.4          & \multicolumn{1}{c|}{68.1 $\pm$ 3.3}          & \multicolumn{1}{c|}{71.2 $\pm$ 2.8}           & 73.0 $\pm$ 0.2          & \multicolumn{1}{c|}{53.2 $\pm$ 6.4}          & \multicolumn{1}{c|}{53.6 $\pm$ 5.6}          & 61.0 $\pm$ 5.3          \\ \hline
Featprop                   & \multicolumn{1}{c|}{29.2 $\pm$ 0.7}          & \multicolumn{1}{c|}{31.2 $\pm$ 4.5}          & 49.5 $\pm$ 3.2          & \multicolumn{1}{c|}{\textbf{69.6 $\pm$ 3.2}} & \multicolumn{1}{c|}{72.8 $\pm$ 2.1}           & 74.5 $\pm$ 3.4          & \multicolumn{1}{c|}{57.2 $\pm$ 2.4}          & \multicolumn{1}{c|}{56.4 $\pm$ 3.2}          & 59.7 $\pm$ 6.5          \\ \hline
GraphPart                  & \multicolumn{1}{c|}{29.7 $\pm$ 4.3}          & \multicolumn{1}{c|}{32.4 $\pm$ 2.3}          & 50.1 $\pm$ 4.8          & \multicolumn{1}{c|}{64.2 $\pm$ 3.6}          & \multicolumn{1}{c|}{70.4 $\pm$ 5.8}           & 72.5 $\pm$ 3.5          & \multicolumn{1}{c|}{53.6 $\pm$ 7.8}          & \multicolumn{1}{c|}{54.2 $\pm$ 0.6}          & 54.2 $\pm$ 3.4          \\ \hline
GraphPartfar               & \multicolumn{1}{c|}{28.2 $\pm$ 0.7}          & \multicolumn{1}{c|}{35.4 $\pm$ 2.9}          & 52.2 $\pm$ 9.8          & \multicolumn{1}{c|}{64.4 $\pm$ 1.5}          & \multicolumn{1}{c|}{69.4 $\pm$ 3.9}           & 72.2 $\pm$ 1.5          & \multicolumn{1}{c|}{52.1 $\pm$ 3.6}          & \multicolumn{1}{c|}{54.0 $\pm$ 6.8}          & 52.2 $\pm$ 2.6          \\ \hline \hline
SPA                        & \multicolumn{1}{c|}{\textbf{32.8 $\pm$ 2.9}} & \multicolumn{1}{c|}{\textbf{37.2 $\pm$ 2.5}} & \textbf{52.8 $\pm$ 1.6} & \multicolumn{1}{c|}{68.8 $\pm$ 1.8}          & \multicolumn{1}{c|}{\textbf{73.2 $\pm$ 0.1}}  & \textbf{75.2 $\pm$ 2.1} & \multicolumn{1}{c|}{\textbf{58.2 $\pm$ 1.5}} & \multicolumn{1}{c|}{\textbf{59.6 $\pm$ 3.4}} & \textbf{61.2 $\pm$ 1.5} \\ \hline \hline
\end{tabular}}\label{tab:resultssage}
\end{table*}
\subsection{Complexity Analysis}
The computational complexity of PageRank is typically $\mathcal{O}(n + m)$ per iteration in a graph $G$ with $n$ nodes and $m$ edges. Additionally, the SCAN algorithm has a complexity of $\mathcal{O}(m \sqrt{m})$. This complexity is primarily dictated by the clustering process involving each edge and its neighboring nodes. While selecting the highest PageRank node in each community incurs additional computational overhead, this is generally less significant compared to the overall complexities of PageRank and SCAN. The overhead primarily depends on the number of communities and the size of the graph. Therefore, the overall performance of this combined approach is influenced by the size and connectivity of the input graph, with larger and more connected graphs incurring higher computational costs. When compared with recent state-of-the-art models like GraphPart and GraphPartFar, the proposed method demonstrates lower computational costs, leading to reduced query times.
To compare the computational complexity of the proposed method, we perform a computational cost experiment against the recent model. We measure the cost of each model in terms of query time which is the time needed from each model to calculate which sample to be labeled by the active learning process. Table \ref{tab:execution time} presents the query times for each method. The proposed method notably reduces query time, with the most significant reduction seen in the Corafull dataset. In the Corafull dataset, SPA reduced the execution time down to 25 seconds, compared to 319 seconds for GraphPart and 397 seconds for GraphPartFar.
\vspace{-8mm}
\begin{table}[h!]
\centering
\caption{Query Time Comparison: Proposed Method vs. State-of-the-Art (Average over 10 Runs, Measured in Seconds}
\scalebox{0.9}{
\begin{tabular}{c|ccc}
\hline \hline
\multirow{2}{*}{Dataset} & \multicolumn{3}{c}{Graph Architecture=GCN}                               \\ \cline{2-4} 
                         & \multicolumn{1}{c|}{GraphPart} & \multicolumn{1}{c|}{GraphPartFar} & SPA  \\ \hline
Citeseer                 & \multicolumn{1}{c|}{10}        & \multicolumn{1}{c|}{20}           & 0.21 \\ \hline
Pubmed                   & \multicolumn{1}{c|}{13.3}      & \multicolumn{1}{c|}{17}           & 4    \\ \hline
Corafull                 & \multicolumn{1}{c|}{316}       & \multicolumn{1}{c|}{397}          & 25   \\ \hline
Wikics                   & \multicolumn{1}{c|}{45}        & \multicolumn{1}{c|}{52}           & 23   \\ \hline
Minesweeper              & \multicolumn{1}{c|}{25}        & \multicolumn{1}{c|}{30}           & 3.2  \\ \hline
Tolokers                 & \multicolumn{1}{c|}{516}       & \multicolumn{1}{c|}{621}          & 115  \\ \hline \hline
\end{tabular}}\label{tab:execution time}
\end{table}

\vspace{-10mm}
\section{Discussion and Conclusion}
This paper introduced a Structural-Clustering based Active Learning (SPA) approach for Graph Neural Networks (GNNs), which combines community detection with the PageRank scoring method. The SPA method strategically prioritizes nodes based on their information content and centrality within the graph's structure, leading to a more representative sample selection and enhancing the robustness of active learning outcomes. This is particularly effective in real-world applications like social network analysis and financial networks, which typically struggle with large amounts of labeled data requirements. SPA's efficiency across varying annotation budgets is an important advantage in scenarios with limited resources for labeling.

Furthermore, SPA integrates the structural clustering abilities of the SCAN algorithm with the PageRank scoring system. SCAN uses both feature and structural information in graphs to identify community-based local structures, while PageRank focuses on the global importance of nodes. The proposed method has demonstrated improved execution times and superior Macro-F1 scores across various datasets, yet it may face potential challenges in extremely large or complex graph structures.

In conclusion, the SPA method represents a substantial advancement in the field of active learning for GNNs. It not only improves performance but also enhances execution efficiency, marking an important step in applying active learning to graph-structured data. While showing promising results in various scenarios, we acknowledge the need for further research in optimizing the method for large-scale or complex graphs. The insights from this research lay the groundwork for future developments in comprehensive and efficient active learning models, catering to a wide range of active learning applications in graph-structured data.

%
%
%
%
\bibliographystyle{splncs04}
\bibliography{references}
\end{document}